\pdfoutput=1
\documentclass[10pt,twocolumn,letterpaper]{article}

\usepackage{cvpr}              

\usepackage{graphicx}
\usepackage{amsmath}
\usepackage{amssymb}
\usepackage{booktabs}
\usepackage{multirow}
\usepackage{algorithm}
\usepackage{algorithmic}
\usepackage{makecell}
\usepackage[accsupp]{axessibility}

\usepackage[pagebackref,breaklinks,colorlinks]{hyperref}

\usepackage[capitalize]{cleveref}
\crefname{section}{Sec.}{Secs.}
\Crefname{section}{Section}{Sections}
\Crefname{table}{Table}{Tables}
\crefname{table}{Tab.}{Tabs.}

\def\cvprPaperID{5540} 
\def\confName{CVPR}
\def\confYear{2022}

\begin{document}

\title{Towards Practical Certifiable Patch Defense with Vision Transformer}

\author{Zhaoyu Chen$^{1}$\footnotemark[1]$\quad$
        Bo Li$^{2}$\footnotemark[1]  \footnotemark[2]$\quad$
        Jianghe Xu$^{2}\quad$
        Shuang Wu$^{2}\quad$
        Shouhong Ding$^{2}\quad$ 
        Wenqiang Zhang$^{1}$\footnotemark[2] \\ 
        $^1$Academy for Engineering and Technology,\ Fudan University$\quad$
        $^2$Tencent Youtu Lab\\
{\tt\small $\{$zhaoyuchen20,wqzhang$\}$@fudan.edu.cn}$\quad$
{ \tt\small$\{$libraboli,jankosxu,calvinwu,ericshding$\}$@tencent.com}
}
\maketitle

\renewcommand{\thefootnote}{\fnsymbol{footnote}} 
\footnotetext[1]{indicates equal contributions.} 
\footnotetext[2]{indicates corresponding author.} 

\begin{abstract}
Patch attacks, one of the most threatening forms of physical attack in adversarial examples, can lead networks to induce misclassification by modifying pixels arbitrarily in a continuous region. Certifiable patch defense can guarantee robustness that the classifier is not affected by patch attacks. Existing certifiable patch defenses sacrifice the clean accuracy of classifiers and only obtain a low certified accuracy on toy datasets. Furthermore, the clean and certified accuracy of these methods is still significantly lower than the accuracy of normal classification networks, which limits their application in practice. To move towards a practical certifiable patch defense, we introduce Vision Transformer (ViT) into the framework of Derandomized Smoothing (DS). Specifically, we propose a progressive smoothed image modeling task to train Vision Transformer, which can capture the more discriminable local context of an image while preserving the global semantic information. For efficient inference and deployment in the real world, we innovatively reconstruct the global self-attention structure of the original ViT into isolated band unit self-attention. On ImageNet, under 2\% area patch attacks our method achieves 41.70\% certified accuracy, a nearly 1-fold increase over the previous best method (26.00\%). Simultaneously, our method achieves 78.58\% clean accuracy, which is quite close to the normal ResNet-101 accuracy. Extensive experiments show that our method obtains state-of-the-art clean and certified accuracy with inferring efficiently on CIFAR-10 and ImageNet.
\end{abstract}

\section{Introduction}
\label{sec:intro}

Despite achieving outstanding performance on various computer vision tasks~\cite{DBLP:conf/aaai/LiSG19,DBLP:conf/ijcai/0061STSS19,DBLP:conf/iccv/LiSLWH19,DBLP:conf/mm/LiSWL19,DBLP:conf/icassp/LiSTH19,DBLP:conf/iccv/TangLZDS21,DBLP:journals/corr/abs-2110-12748,zhao2020efficient,kong2021reflash,Kong_2021_CVPR,liu2022efficient}, deep neural networks (DNNs) are vulnerable and susceptive to adversarial examples~\cite{fgsm,pgd,cmua,rpattack}, which are attached to a perturbation on images by adversaries~\cite{L_BFGS}. Patch attack is one of the most threatening forms of adversarial examples, which can modify pixels arbitrarily in a continuous region, and can implement physical attacks on autonomous systems via their perception component. For example, putting stickers on traffic signals can make the model misprediction~\cite{traffic_sign}.

While several practical patch defenses are proposed~\cite{digital_watermark, LGS}, they only obtain robustness against known attacks but not against more powerful attacks that may be developed in the future~\cite{ibp,adaptiveattack}. Therefore, we focus on certifiable defense against patch attacks in this paper, which allows guaranteed robustness against all possible attacks for the given threat model. In recent years, this community has received great attention. For example, Chiang et al. propose the first certifiable defense against patch attacks by extending interval bound propagation (IBP)~\cite{ibp} on CIFAR10. Then later work introduces small receptive fields or randomized smoothing to improve certification on CIFAR10 and scale to ImageNet.

However, the accuracy gap between certifiable patch defense and normal model limits the practical application of these defense methods. For example, PatchGuard~\cite{PatchGuard} can achieve 84.7\% clean accuracy and 57.7\% certified accuracy under 4$\times$4 patches on CIFAR10. However, when PatchGuard~\cite{PatchGuard} is extended to large-scale datasets, such as ImageNet, it can only obtain 54.6\% clean accuracy and 26.0\% certified accuracy under 2\% patches, which is much lower than the normal ResNet-50~\cite{resnet} (76.2\%). 
Consequently, a breakthrough is needed urgently to narrow the gap and move towards practical certifiable patch defenses.

Recently, transformer~\cite{transformer} has achieved significant success in speech recognition and natural language processing. Inspired by this, Vision Transformer (ViT)~\cite{vit} has been proposed and obtain potential performance in computer vision, such as image classification~\cite{vit, beit}, objects detection~\cite{dpt} and semantic segmentation~\cite{beit}. ViT models the context between different patches and obtains long-range dependencies by self-attention. Compared with convolutional neural networks (CNNs), ViT has achieved promising performance, which has the potential to improve certification. Furthermore, Derandomize smoothing (DS)~\cite{ds} is a classic certifiable patch defense based on randomized smoothing robustness schemes and provides high confident certified robustness by structured ablation. It can also be generalized to other network architectures. Therefore, integrating ViT into DS is a potential certifiable patch defense. 
However, direct replacing the CNN structure in DS with ViT leads to trivial results:
(1) the accuracy is still lower than normal classification networks; (2) the excessive inference time limits the application of the method in practice. 

To address these issues and move towards practical certifiable patch defense, we propose an efficient certifiable patch defense with ViT to improve accuracy and inference efficiency. 
First, we introduce a progressive smoothed image modeling task to train ViT. Specifically, the training objective is to gradually recover the original image tokens based on the smoothed image bands.
By gradually reconstructing, the base classifier can explicitly capture the local context of an image while preserving the global semantic information. Consequently, more discriminative local representations can be obtained through very limited image information (a thin smoothed image band), which improves the performance of the base classifier. 
Then, we renovate the global self-attention structure of the original ViT into the isolated band-unit self-attention. The input image is divided into bands and the self-attention in each band-like unit is calculated separately, which provides the feasibility for the parallel calculation of multiple bands. Finally, our method achieves 78.58\% clean accuracy on ImageNet and 41.70\% certified accuracy within efficient inference under 2\% area patch attacks. The clean accuracy is quite close to the normal ResNet-101 accuracy. Extensive experiments demonstrate that our method obtains state-of-the-art clean and certified accuracy with inferring efficiently on CIFAR-10 and ImageNet.
Our major contributions are as follows:
\begin{itemize}
    \item We introduce ViT into certifiable patch defense and propose a progressive smoothed image modeling task, which lets the model capture more discriminable local context of an image while preserving the global semantic information.
    \item We renovate the global self-attention structure of the ViT into the isolated band-unit self-attention, which considerably accelerates inference.
    \item Experiments show that our method obtains state-of-the-art clean and certified accuracy with inferring efficiently on CIFAR-10 and ImageNet. Additionally, our method achieves 78.58\% clean accuracy on ImageNet and 41.70\% certified accuracy in efficient inference under 2\% area patch attacks. The clean accuracy is quite close to the normal ResNet-101 accuracy.
\end{itemize}

\begin{figure*}[t]
\begin{minipage}[b]{1\linewidth}
  \centering
  \centerline{\includegraphics[width=15cm]{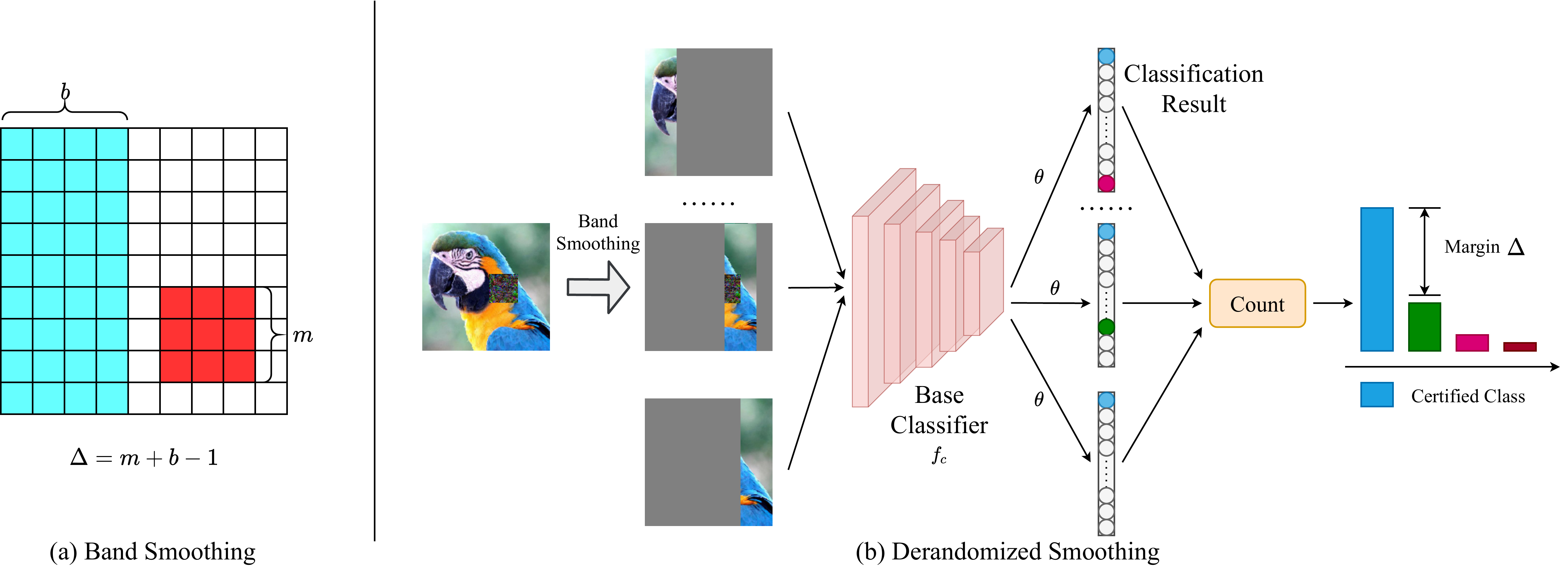}}
\end{minipage}
\caption{Introduction of Derandomized Smoothing (DS). The red patch represents the adversarial patch and the blue band represents the retained image after the band smoothing in (a). (b) describes the pipeline of DS. First, DS smoothes the image in the band smoothing and obtains the smoothed images from different positions. Then the smoothed images are fed into the base classifier $f_c$ and we obtain the classification result by the threshold $\theta$. Finally, DS counts the result and applies Equation~\ref{eq:certification} to judge whether the image is certified.}
\label{fig:ds}
\vspace{-0.3cm}
\end{figure*}

\section{Related Work}
\subsection{Patch Attacks}
Patch attacks are one of the most threatening forms of physical attacks, in which adversaries can arbitrarily modify pixels within the small continuous region. GAP~\cite{adv_patch} first creates universal, robust, targeted adversarial image patches in the real world and causes a classifier to output any target class. Then LaVAN~\cite{lavan} shows that it is possible to learn visible and localized adversarial patches that cover only 2\% of the pixels in the image and cause image classifiers to misclassify to arbitrary labels in the digital domain. Because patch attack can be implemented in the form of stickers in the physical world, it brings great harm to vision systems, such as object detection~\cite{camouflage_detect, cloak} and visual tracking~\cite{mtd}.

\subsection{Certifiable Patch Defense}
Several practical patch defenses were proposed such as digital watermark~\cite{digital_watermark} and local gradient smoothing~\cite{LGS}. However, Chiang et al.~\cite{ibp} demonstrate that these defenses can be easily broken by white-box attacks which account for the pre-processing steps in the optimization procedure. It means that practical patch defenses only obtain robustness against known attacks but not against more powerful attacks that may be developed in the future.

Therefore, it is important to have guarantees of robustness in face of the worst-case adversarial patches. Recent work~\cite{DBLP:conf/mm/LiXWDLH21} focuses on certified defenses against patch attacks, which allow guaranteed robustness against all possible attacks for the given threat model. Chiang et al.~\cite{ibp} propose the first certifiable defense against patch attacks by extending interval bound propagation (IBP) on MNIST and CIFAR10. However, it is hard to scale to the ImageNet. Levine et al. propose Derandomized Smoothing (DS)~\cite{ds}, which trains a base classifier by smoothed images and a majority vote determines the final classification. This method provides significant accuracy improvement when compared to IBP on ImageNet but its inference is computationally expensive. Some work is based on using CNNs with the small receptive field, such as Clipped BagNet (CBN)~\cite{cbn}, Patchguard~\cite{PatchGuard} and BagCert~\cite{bagcert}. 

\subsection{Vision Transformer}
Transformer~\cite{transformer} is the mainstream method in the natural language processing field, which captures long-range dependencies through self-attention and achieves state-of-the-art performance. Vision Transformer (ViT)~\cite{vit} is the first work to achieve comparable results with traditional CNN architectures constructed only by self-attention blocks. It divides the image into a sequence of fixed-size patches and models the context between different patches and obtains long-range dependencies by multi-head self-attention.

The accuracy and certified robustness of the existing certifiable patch defenses are still not enough to be applied in practice. 
We introduce ViT into certifiable patch defense with the progressive smoothed image modeling task. With isolated band unit self-attention, our method achieves significant improvements in accuracy and inference efficiency, which enables practical certifiable patch defense.

\begin{algorithm}[t]
  \caption{Progressive Smoothed Image Modeling Task}
  \label{alg:training}
  \textbf{Input}: the image $x$, the label $Y$, the tokenizer $Z$, transformer encoder $f$, weighting factor $\lambda$, MLP head $mlp$ and the number of stages $N_s$
  \\
  \textbf{Output}: $f$, $mlp$
  \begin{algorithmic}[1]
  \FOR{$i \in [1,\ N_s]$}
    \STATE Smooth images $x$ and obtain smoothed images $x_s$
    \STATE Determine expected reconstructed images $x_e$
    \STATE Calculate visual tokens $z$ with $x$ via the tokenizer $Z$
    \STATE Calculate the output representation $H_O$ with $x_e$ via the transformer encoder $f$
    \STATE Calculate logits $l$ with $H_O$ via the MLP head $mlp$
    \STATE Select the reconstructed tokens $H_R$ and corresponding visual tokens $Z_R$ by Equation~\ref{hr} and Equation~\ref{zr}
    
    \STATE Calculate the loss $L$ by Equation~\ref{ce} or Equation~\ref{l2}
    \STATE Update $f$ and $mlp$ through $L$ backward
  \ENDFOR
    \RETURN $f$, $mlp$
  \end{algorithmic}
\end{algorithm}

\section{Method}
In this section, we first review the certified mechanism of DS. Second, we propose a progressive smoothed image modeling task to help ViT capture the more discriminable local context of an image while preserving the global semantic information. Finally, we propose the isolated band unit self-attention to accelerate inference and move towards practical certifiable patch defense.

\subsection{Preliminaries}
Smoothing in derandomized smoothing means to keep a part of the continuous image and smooth other parts of the image. For example, band smoothing means smoothing the entire image except for a band of a fixed-width $b$, as shown in Figure~\ref{fig:ds} (a). DS trains the base classifier with smoothed images. For an input image $x \in R^{c \times h\times w}$, let the base classifier be expressed as $f_c(x,b,p,\theta)$, where $x$ is the input images, $b$ is the width of the band, $p$ is the position of the retained band, $\theta$ is the threshold for voting and $c$ is the class label. For each class $c$, $f_c(x,b,p,\theta)$ is $1$ if its logits of the class $c$ is greater than the threshold $\theta$, otherwise it is $0$. To calculate certified robustness, DS counts the number of bands on which the base classifier is applied to each class.
\begin{equation}
    \forall c,\quad n_c(p)=\sum_{p=1}^w f_c(x,b,p,\theta).
\end{equation}

An image is certified only if the statistics of the highest class (e.g. the label $c$) are greater than a margin than the next highest class $c'$. The shape of the adversarial patch is supposed to be $m \times m$. The number of intersections between the band and this patch is at most $\Delta=m+b-1$. Therefore, an image is certified when $\Delta$ satisfies the following conditions:
\begin{equation}
    \label{eq:certification}
    n_c(x)> \max_{c' \neq c} n_{c'}(x) + 2\Delta.
\end{equation}

When the threshold $\theta$ is determined, the highest class has been guaranteed not to be affected by the adversarial patch. Therefore, we define \textbf{clean accuracy} as the accuracy that the classification is correct after voting. \textbf{Certified accuracy} is the accuracy that the classification is correct and $\Delta$ satisfies Equation~\ref{eq:certification} after voting.

\begin{figure*}[tb]
\centering
\centerline{\includegraphics[width=17cm]{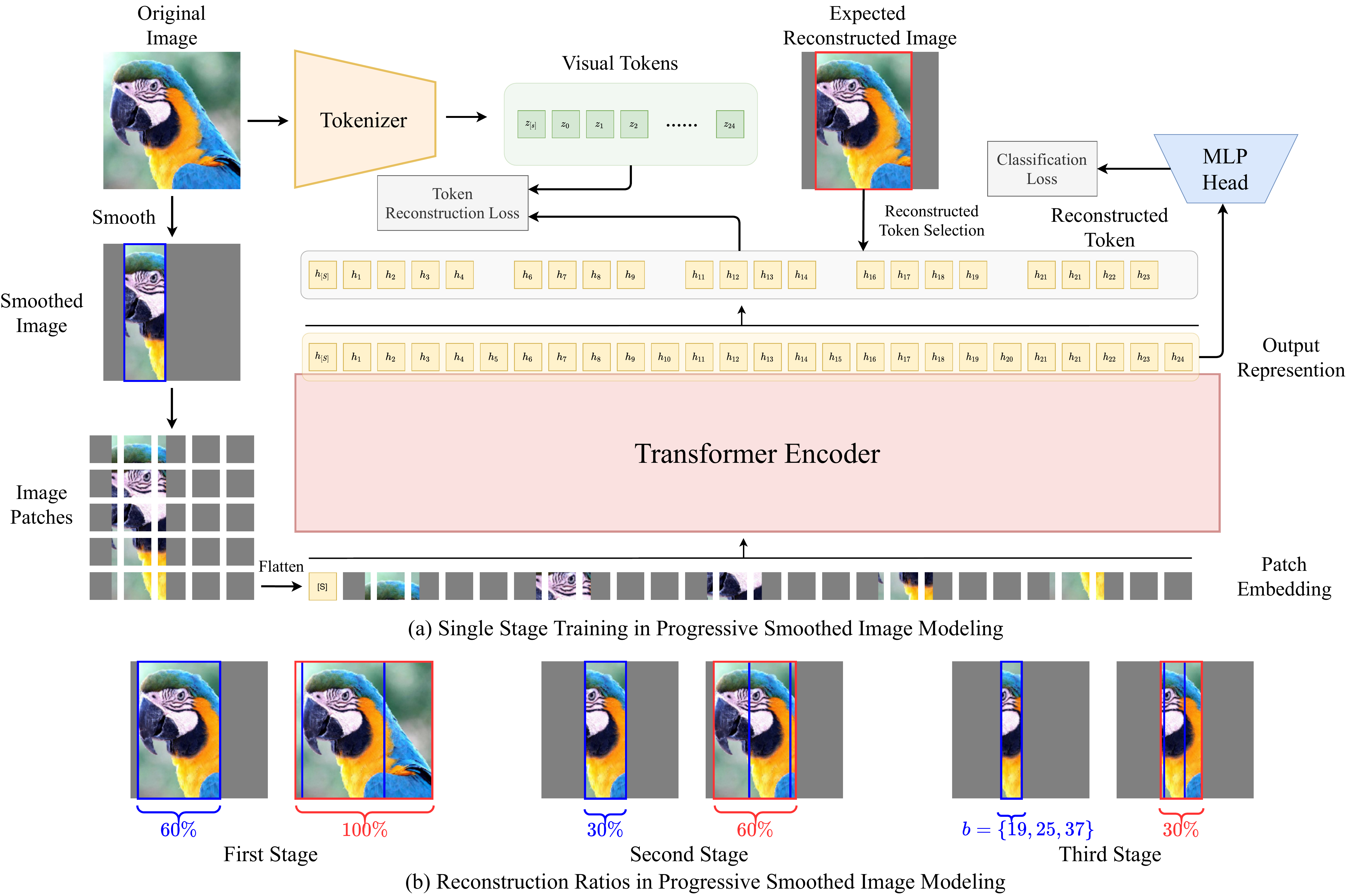}}
\caption{Introduction of Progressive Smoothed Image Modeling. (a) describes a single stage smooth training in progressive smoothed image modeling. We expect smoothed images to be reconstructed as expected reconstruction images. (b) describes reconstructed ratios in multi-stage training. The blue boxes represent the band after smoothing and the red boxes represent the expected reconstructed band.}
\label{fig:training}
\vspace{-0.3cm}
\end{figure*}

\subsection{Progressive Smoothed Image Modeling}
Since the base classifier can only use very limited information (such as bands), we need the base classifier to have the ability to better capture discriminable features. Specifically, in natural language processing, the masked language modeling (MLM) training paradigm (like BERT~\cite{bert}) has been proved to be effective in learning more discriminative features and improving the performance of the model. Inspired by MLM, we propose a smoothed image modeling task to train ViT.

However, unlike languages which are human-created signals with dense semantic correlations between words, 
as natural signals, the visual content of different parts in an image has a high degree of freedom. 
Therefore, it is very difficult to use a band with a width of $b$ ($b \ll h, w$) to recover the full-scale image tokens in one stage.
Hence, we use a multistage smoothed image modeling task called progressive smoothed image modeling to train the base classifier, as shown in Figure~\ref{fig:training}.
By gradually reconstructing the smoothed image parts, the base classifier can explicitly capture the local context of an image while preserving the global semantic information. Consequently, more discriminative local representations can be obtained through very limited image information, which improves the performance of the base classifier. 

In ViT, an image is spilt into a sequence of patches as inputs. Formally, we need to flatten a image $x\in R^{c \times h \times w}$ into $(N=hw/p^2)$ patches $x^p \in R^{N \times p^2c}$, where the shape of the image $x$ is $(h,w)$, the number of channels is $c$, and $(p, p)$ is the shape of patches (e.g. $p=16$). 
The patches $\{x^p_i\}^N_{i=1}$ are projected to obtain the patch embeddings $\{Ex^p_i\}^N_{i=1}$, where $E \in R^{p^2c \times d}$ and $d$ is the embedding dimension.
Like Bert~\cite{bert}, we concatenate class token $E_{[s]}$  to patch embeddings $Ex^p_i$. 
Simultaneously, in order to encode position information, we need to add 1D learnable position embeddings $E_{pos}$ to patch embeddings $Ex^p_i$. Then, the input vector $H_I=[E_{[s]}, Ex^p_i,...,Ex^p_N]+E_{pos}$ is fed to transformer and $H_O=[h_{[s]}, h_i,...,h_N]$ is used as the output representation for the image patches with respect to $x$.

Here, we first introduce single stage training in progressive smoothed image modeling, as illustrated in Figure~\ref{fig:training} (a). BERT-based training has been explored in vision tasks. The difficulty is that it is non-trivial to recover tokens in computer vision. 
For accelerating convergence, we introduce a tokenizer as the supervision of reconstruction. There are two types of supervision for reconstruction: VAE and distillation. 
For VAE, we use pre-trained VAE\cite{dall_e} for supervision.
For distillation, we use the output of pre-trained ViT~\cite{vit} for supervision.
As given a smoothed image $x_s$, we split it into $N$ image patches $\{{x_s}^p_i\}^N_{i=1}$ and obtain $N$ visual tokens $\{z_i\}^N_{i=1}$. 
Centered on the band of $x_s$, we select the reconstructed band and generate the expected reconstructed image.
According to the expected reconstructed image, we have a reconstructed token selection to obtain reconstructed tokens. The band in the expected reconstructed image produces a band mask $\{M^b_i\}^N_{i=1}$ corresponding to the patch $x^p_i$ which needs constructing. 
Hence, the reconstructed tokens and corresponding visual tokens are rephrased as:
\begin{equation}
    H_R=\{h_i: M^b_i=1\}^N_{i=1}.
    \label{hr}
\end{equation}
\begin{equation}
    Z_R=\{z_i: M^b_i=1\}^N_{i=1}.
    \label{zr}
\end{equation}
\indent The objective of Smoothed Training is to simultaneously minimize the classification loss and token reconstruction loss. For VAE, the total loss can be expressed as:
\begin{equation}
    \min \underbrace{CE(l,\ Y)}_{\mathrm{classification\ loss}} + \lambda \cdot \underbrace{CE(Z_R,\ H_R)}_{\mathrm{token\ reconstruction\ loss}}.
    \label{ce}
\end{equation}
\indent For distillation, the total loss can be expressed as:
\begin{equation}
    \min \underbrace{CE(l,\ Y)}_{\mathrm{classification\ loss}} + \lambda \cdot \underbrace{||Z_R-H_R||_2}_{\mathrm{token\ reconstruction\ loss}}.
    \label{l2}
\end{equation}

Here, $l$ is the output logits after passing MLP head and $Y$ is the label of $x$. $\lambda=1000$ balances the gradients between token reconstruction loss and classification loss.

Figure~\ref{fig:training} (b) shows the reconstruction ratio varies within each stage. 
The blue boxes represent the band after smoothing and the red boxes represent the expected reconstructed band. In the first stage, we randomly smooth the approximately 40\% image. The remaining 60\% of the images are used to reconstruct the whole image. In the second stage, we smooth away 70\% of the images and utilize the remaining 30\% of the bands, to reconstruct 60\% of the patches within a neighborhood centered on the 30\% band, including the 30\% band. In the last stage, only the band with width $b$ is reserved, and all other parts are smoothed. The band with width $b$ is used to reconstruct 30\% of the patches in the neighborhood centered on the band. Our method greatly narrows the accuracy gap between DS and normal model by progressive smoothed image modeling, making it possible to achieve certifiable patch defense in practice.

\begin{figure}[tb]
\centering
\centerline{\includegraphics[width=8.5cm]{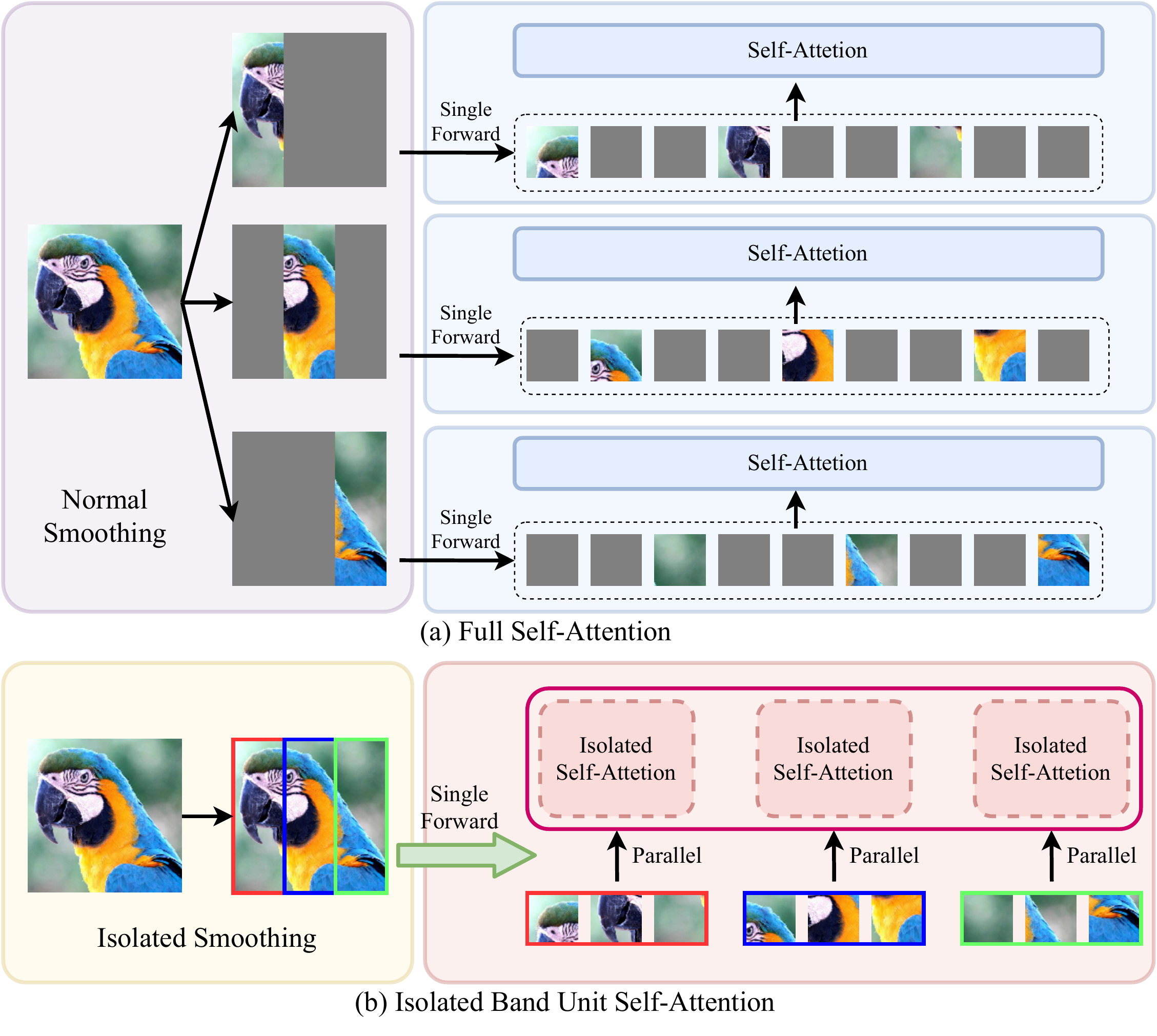}}
\caption{Introduction of Isolated Band Unit Self-attention. (a) describes the normal training that smoothed parts are redundant and unnecessary to calculate. (b) introduces the isolated band unit self-attention that smoothed parts are dropped and self-attention is only calculated within parallel windows.}
\label{fig:attention}
\vspace{-0.3cm}
\end{figure}

\subsection{Isolated Band Unit Self-attention}
To achieve practical certifiable patch defense, high accuracy needs to be combined with efficient inference. With progressive smoothed image modeling, DS improves the clean and certified accuracy, however, it requires hundreds of inferences for various smoothed images, which limits its application in practice.

Smoothed parts of the smoothed image introduce redundant information and invalid calculation. As shown in Figure~\ref{fig:attention} (a), normal smoothing utilizes the whole smoothed image but its calculation is unnecessary for smoothed parts. The long-range dependencies of ViT use this redundant information, which harms their accuracy and introduces extra calculation cost. Furthermore, it is inefficient to calculate one forward calculation for every smoothed image.

Therefore, we innovatively renovate the global self-attention structure of the original ViT into isolated band unit self-attention. Specifically, the input image is divided into bands by sliding windows, and the self-attention in each band-like unit is calculated separately, which provides the feasibility for the parallel calculation of multiple bands.
As shown in Figure~\ref{fig:attention} (b), we choose patches of each band by parallel sliding windows and infer multiple bands within one forward calculation. In the isolated band unit self-attention, a window is a band and isolated self-attention is only  calculated within the window. 

Compared to normal smoothing, ViT splits $x_s$ into $hw/p^2$ patches and we only select the $N=hb/p^2$ patches within windows for fine-tuning and inferring. The time complexity of the whole image input of self-attention operation is $O\left(N^2 d+N d^2 \right)$, where the first term is the complexity of attention operation and another is the complexity of fully-connected operation. Compared with the input of the whole image, the isolated self-attention can reduce the amount of calculation to $\frac{1}{w/b}$ of the former. Furthermore, it has $\lceil \frac{w}{b} \rceil$ adjacent windows for inference at the same time, where the shape of windows is $(h, b)$. Therefore, we change the original forward calculation from $w$ times to $b\ (w\gg b)$ times. Therefore, it is possible to deploy a certifiable patch defense in real systems through efficient inference.

\begin{table}[t]
\caption{The parameters of models on ImageNet.}
\label{tab:paras}
\centering
\scalebox{0.78}{
\begin{tabular}{@{}ccccccc@{}}
\toprule
Model      & BagNet33 & ViT-s & ResNet50 & ResNext101 & ViT-B   \\ \midrule
Parameters & 18M      & 22M   & 26M      & 88.79M     & 86M \\ \bottomrule
\end{tabular}
}
\centering
\caption{Clean and certified accuracy compared with state-of-the-art certifiable patch defenses on CIFAR10.}
\label{tab:cifar10}
\scalebox{0.85}{
\begin{tabular}{@{}cc|c|cc@{}}
\toprule
\multicolumn{2}{c|}{\multirow{2}{*}{Method}} & \multirow{2}{*}{\begin{tabular}[c]{@{}c@{}}Clean\\ Accuracy (\%)\end{tabular}} & \multicolumn{2}{c}{Certified Accuracy (\%)} \\ \cmidrule(l){4-5} 
\multicolumn{2}{c|}{} &  & \multicolumn{1}{c|}{$\quad 2\times 2\quad$} & \multicolumn{1}{c}{$4 \times 4$} \\ \midrule
\multicolumn{1}{c|}{\multirow{4}{*}{Baseline}} & CBN & 84.20 & \multicolumn{1}{c|}{44.20} & \multicolumn{1}{c}{9.30} \\ \cmidrule(l){2-5} 
\multicolumn{1}{c|}{} & DS & 83.90 & \multicolumn{1}{c|}{68.90} & \multicolumn{1}{c}{56.20} \\ \cmidrule(l){2-5} 
\multicolumn{1}{c|}{} & PG & 84.70 & \multicolumn{1}{c|}{69.20} & \multicolumn{1}{c}{57.70} \\ \cmidrule(l){2-5} 
\multicolumn{1}{c|}{} & BagCert & 86.00 & \multicolumn{1}{c|}{73.33} & \multicolumn{1}{c}{64.90} \\ \midrule
\multicolumn{1}{c|}{\multirow{5}{*}{\makecell[c]{Smooth\\Model}}} & ViT-S & 80.40 & \multicolumn{1}{c|}{61.50} & \multicolumn{1}{c}{51.78} \\ \cmidrule(l){2-5} 
\multicolumn{1}{c|}{} & ECViT-S & 87.56 & \multicolumn{1}{c|}{73.82} & \multicolumn{1}{c}{65.10} \\ \cmidrule(l){2-5} 
\multicolumn{1}{c|}{} & ResNext101 & 85.34 & \multicolumn{1}{c|}{69.32} & \multicolumn{1}{c}{60.68} \\ \cmidrule(l){2-5} 
\multicolumn{1}{c|}{} & ViT-B & 91.28 & \multicolumn{1}{c|}{78.10} & \multicolumn{1}{c}{70.78} \\ \cmidrule(l){2-5} 
\multicolumn{1}{c|}{} & ECViT-B & \textbf{93.48} & \multicolumn{1}{c|}{\textbf{82.80}} & \multicolumn{1}{c}{\textbf{76.38}} \\ \bottomrule
\end{tabular}
}
\caption{Clean and certified accuracy compared with state-of-the-art certifiable patch defenses on ILSVRC2012.}
\scalebox{0.81}{
\centering
\label{tab:imagenet}
\begin{tabular}{@{}c|c|ccc@{}}
\toprule
\multirow{2}{*}{Method} & \multirow{2}{*}{\begin{tabular}[c]{@{}c@{}}Clean\\ Accuracy (\%)\end{tabular}} & \multicolumn{3}{c}{Certified Accuracy (\%)} \\ \cmidrule(l){3-5} 
 &  & \multicolumn{1}{c|}{1\% pixels} & \multicolumn{1}{c|}{2\% pixels} & 3\% pixels \\ \midrule
CBN & 49.50 & \multicolumn{1}{c|}{13.40} & \multicolumn{1}{c|}{7.10} & 3.10 \\ \midrule
PG (1\%) & 55.10 & \multicolumn{1}{c|}{32.30} & \multicolumn{1}{c|}{-} & - \\ \midrule
PG (2\%) & 54.60 & \multicolumn{1}{c|}{26.00} & \multicolumn{1}{c|}{26.00} & - \\ \midrule
PG (3\%) & 54.10 & \multicolumn{1}{c|}{19.70} & \multicolumn{1}{c|}{19.70} & 19.70 \\ \midrule
DS & 64.67 & \multicolumn{1}{c|}{30.14} & \multicolumn{1}{c|}{24.70} & 20.88 \\ \midrule
BagCert & 46.00 & \multicolumn{1}{c|}{-} & \multicolumn{1}{c|}{23.00} & - \\ \midrule
ECViT-S(b=37) & 69.88 & \multicolumn{1}{c|}{35.03} & \multicolumn{1}{c|}{29.74} & 25.74 \\ \midrule
ECViT-B(b=37) & \textbf{78.58} & \multicolumn{1}{c|}{\textbf{47.39}} & \multicolumn{1}{c|}{\textbf{41.70}} & \textbf{37.26} \\ \bottomrule
\end{tabular}
}
\end{table}

\begin{table}[t]
\caption{Clean and certified accuracy compared with smooth models on ILSVRC2012.}
\label{tab:smoothmodel}
\scalebox{0.67}{
\begin{tabular}{@{}c|c|ccc|c@{}}
\toprule
\multirow{2}{*}{Method} & \multirow{2}{*}{\begin{tabular}[c]{@{}c@{}}Clean\\ Accuracy (\%)\end{tabular}} & \multicolumn{3}{c|}{Certified Accuracy (\%)} & \multirow{2}{*}{\makecell[c]{Inference\\Time (s)}} \\ \cmidrule(lr){3-5}
 &  & \multicolumn{1}{c|}{1\% pixels} & \multicolumn{1}{c|}{2\% pixels} & 3\% pixels &  \\ \midrule
ResNet50(b=19) & 62.03 & \multicolumn{1}{c|}{29.03} & \multicolumn{1}{c|}{23.53} & 19.77 & \multirow{3}{*}{69.00} \\ \cmidrule(r){1-5}
ResNet50(b=25) & 64.67 & \multicolumn{1}{c|}{30.14} & \multicolumn{1}{c|}{24.70} & 20.88 &  \\ \cmidrule(r){1-5}
ResNet50(b=37) & 67.60 & \multicolumn{1}{c|}{27.15} & \multicolumn{1}{c|}{21.86} & 18.14 &  \\ \midrule
ViT-S(b=19) & 63.88 & \multicolumn{1}{c|}{33.08} & \multicolumn{1}{c|}{27.78} & 23.84 & \multirow{3}{*}{87.13} \\ \cmidrule(r){1-5}
ViT-S(b=25) & 66.49 & \multicolumn{1}{c|}{33.90} & \multicolumn{1}{c|}{28.59} & 24.57 &  \\ \cmidrule(r){1-5}
ViT-S(b=37) & 69.01 & \multicolumn{1}{c|}{33.37} & \multicolumn{1}{c|}{28.07} & 24.15 &  \\ \midrule
ECViT-S(b=19) & 64.69 & \multicolumn{1}{c|}{34.38} & \multicolumn{1}{c|}{28.85} & 24.74 & 9.66 \\ \midrule
ECViT-S(b=25) & 67.14 & \multicolumn{1}{c|}{\textbf{35.57}} & \multicolumn{1}{c|}{\textbf{30.06}} & \textbf{25.98} & 13.20 \\ \midrule
ECViT-S(b=37) & \textbf{69.88} & \multicolumn{1}{c|}{35.03} & \multicolumn{1}{c|}{29.74} & 25.74 & 23.54 \\ \midrule\midrule
ResNext101(b=19) & 69.36 & \multicolumn{1}{c|}{40.74} & \multicolumn{1}{c|}{34.97} & 30.58 & \multirow{3}{*}{567.75} \\ \cmidrule(r){1-5}
ResNext101(b=25) & 71.96 & \multicolumn{1}{c|}{41.86} & \multicolumn{1}{c|}{36.03} & 32.17 &  \\ \cmidrule(r){1-5}
ResNext101(b=37) & 74.89 & \multicolumn{1}{c|}{42.79} & \multicolumn{1}{c|}{36.69} & 33.24 &  \\ \midrule
ViT-B(b=19) & 66.92 & \multicolumn{1}{c|}{34.65} & \multicolumn{1}{c|}{28.71} & 24.80 & \multirow{3}{*}{136.75} \\ \cmidrule(r){1-5}
ViT-B(b=25) & 70.57 & \multicolumn{1}{c|}{36.72} & \multicolumn{1}{c|}{31.13} & 26.80 &  \\ \cmidrule(r){1-5}
ViT-B(b=37) & 74.68 & \multicolumn{1}{c|}{37.61} & \multicolumn{1}{c|}{31.88} & 27.44 &  \\ \midrule
ECViT-B(b=19) & 73.49 & \multicolumn{1}{c|}{46.83} & \multicolumn{1}{c|}{40.72} & 36.29 & 16.63 \\ \midrule
ECViT-B(b=25) & 75.30 & \multicolumn{1}{c|}{46.56} & \multicolumn{1}{c|}{40.79} & 36.21 & 22.72 \\ \midrule
ECViT-B(b=37) & \textbf{78.58} & \multicolumn{1}{c|}{\textbf{47.39}} & \multicolumn{1}{c|}{\textbf{41.70}} & \textbf{37.26} & 40.50 \\ \bottomrule
\end{tabular}
}
\end{table}

\section{Experiments}
We conduct extensive experiments on CIFAR10~\cite{cifar10} and ImageNet~\cite{imagenet}. First, we compare our method with the state-of-the-art certifiable patch defenses on both datasets. In addition, we conduct ablation studies to investigate the factors that affect clean and certified accuracy.

\subsection{Experimental Setup}
In our experiments, we use Pytorch for the implementation and train on NVIDIA Tesla V100 GPUs. We choose different networks as base classifiers, such as ResNet50~\cite{resnet}, ResNext101-32x8d (ResNext101)~\cite{resnext}, ViT-S/16-224 (ViT-S) and ViT-B/16-224 (ViT-B)~\cite{vit}. In the smooth model, the methods directly apply the corresponding backbone into DS and are all fine-tuned from the ImageNet pre-trained model~\cite{timm}. The parameters of models on ImageNet are shown in Table~\ref{tab:paras}. We train the models in 120 epochs on ImageNet and 600 epochs on CIFAR10.

The parameters of our Efficient Certifiable ViT (ECViT) is the same as the ViT. We train ECViT from the same ViT pre-trained models. For exmaple, ECViT-B is based on ViT-B and ECViT-S is based on ViT-S. For CIFAR10, we train ECViT 150 epochs for each stage during the progressive smoothed image modeling task and fine-tune 150 epochs in isolated band unit self-attention. For ImageNet, we train ECViT 30 epochs for each stage during the progressive smoothed image modeling task and fine-tune 30 epochs in isolated band unit self-attention.  

We report clean and certified accuracy compared ECViT with Interval Bound Propagation (IBP)~\cite{ibp}, Derandomized Smoothing (DS)~\cite{ds}, Clipped BagNet (CBN)~\cite{cbn}, Patchguard~\cite{PatchGuard} (PG) and BagCert~\cite{bagcert}. Here, DS is based on the band smoothing and Patchguard is based on the mask BagNet~\cite{bagnet}. For each stage and fine-tuning in the progressive smoothed image modeling or smoothed models, we set the optimizer to AdamW, the loss function to be a cross-entropy loss, the batch size to 512, the warm-up epoch to 5, the learning rate to be 2e-5, the threshold $\theta$ to $0.2$, and the weight decay to be 1e-8. ECViT is trained on CIFAR10 and ImageNet for a total of 600 and 120 epochs, consistent with smoothed models.

\begin{table}[t]
\centering
\caption{Ablation study of training for different stages and tokenizers on ILSVRC2012 .}
\label{tab:stage_imageNet}
\scalebox{0.7}{
\begin{tabular}{@{}c|c|c|c|ccc@{}}
\toprule
\multirow{2}{*}{Network} & \multirow{2}{*}{Band Size} & \multirow{2}{*}{Stages} & \multirow{2}{*}{\makecell[c]{Clean\\Accuracy (\%)}} & \multicolumn{3}{c}{Certified Accuracy (\%)} \\ \cmidrule(l){5-7} 
 &  &  &  & \multicolumn{1}{c|}{1\%} & \multicolumn{1}{c|}{2\%} & 3\% \\ \midrule
\multirow{9}{*}{Distllation} & \multirow{3}{*}{b=19} & one\_stage & 72.01 & \multicolumn{1}{c|}{43.49} & \multicolumn{1}{c|}{37.64} & 33.31 \\ \cmidrule(l){3-7} 
 &  & two\_stage & 72.78 & \multicolumn{1}{c|}{44.87} & \multicolumn{1}{c|}{39.01} & 34.63 \\ \cmidrule(l){3-7} 
 &  & three\_stage & 72.93 & \multicolumn{1}{c|}{45.78} & \multicolumn{1}{c|}{40.03} & 35.61 \\ \cmidrule(l){2-7} 
 & \multirow{3}{*}{b=25} & one\_stage & 74.48 & \multicolumn{1}{c|}{44.14} & \multicolumn{1}{c|}{38.39} & 33.87 \\ \cmidrule(l){3-7} 
 &  & two\_stage & 75.09 & \multicolumn{1}{c|}{45.63} & \multicolumn{1}{c|}{39.91} & 35.54 \\ \cmidrule(l){3-7} 
 &  & three\_stage & 75.12 & \multicolumn{1}{c|}{46.63} & \multicolumn{1}{c|}{40.93} & 36.49 \\ \cmidrule(l){2-7} 
 & \multirow{3}{*}{b=37} & one\_stage & 78.05 & \multicolumn{1}{c|}{44.89} & \multicolumn{1}{c|}{39.24} & 34.60 \\ \cmidrule(l){3-7} 
 &  & two\_stage & 77.98 & \multicolumn{1}{c|}{45.54} & \multicolumn{1}{c|}{39.94} & 35.45 \\ \cmidrule(l){3-7} 
 &  & three\_stage & 78.05 & \multicolumn{1}{c|}{46.46} & \multicolumn{1}{c|}{40.84} & 36.39 \\ \midrule
\multirow{9}{*}{VAE} & \multirow{3}{*}{b=19} & one\_stage & 72.60 & \multicolumn{1}{c|}{43.91} & \multicolumn{1}{c|}{38.00} & 33.60 \\ \cmidrule(l){3-7} 
 &  & two\_stage & 73.30 & \multicolumn{1}{c|}{45.50} & \multicolumn{1}{c|}{39.69} & 35.12 \\ \cmidrule(l){3-7} 
 &  & three\_stage & 73.49 & \multicolumn{1}{c|}{46.83} & \multicolumn{1}{c|}{40.72} & 36.29 \\ \cmidrule(l){2-7} 
 & \multirow{3}{*}{b=25} & one\_stage & 75.15 & \multicolumn{1}{c|}{45.50} & \multicolumn{1}{c|}{39.56} & 35.18 \\ \cmidrule(l){3-7} 
 &  & two\_stage & 75.48 & \multicolumn{1}{c|}{46.34} & \multicolumn{1}{c|}{40.49} & 35.99 \\ \cmidrule(l){3-7} 
 &  & three\_stage & 75.30 & \multicolumn{1}{c|}{46.56} & \multicolumn{1}{c|}{40.79} & 36.21 \\ \cmidrule(l){2-7} 
 & \multirow{3}{*}{b=37} & one\_stage & 77.95 & \multicolumn{1}{c|}{44.49} & \multicolumn{1}{c|}{38.78} & 34.28 \\ \cmidrule(l){3-7} 
 &  & two\_stage & 78.40 & \multicolumn{1}{c|}{46.53} & \multicolumn{1}{c|}{40.73} & 36.36 \\ \cmidrule(l){3-7} 
 &  & three\_stage & 78.58 & \multicolumn{1}{c|}{47.36} & \multicolumn{1}{c|}{41.70} & 37.26 \\ \bottomrule
\end{tabular}
}
\end{table}

\subsection{Certification on CIFAR10}
Following the setting of the previous work~\cite{ds}, we evaluate clean and certified accuracy on 5,000 images of the CIFAR10 validation set.  We select two patch sizes, including $2\times 2$ and $4\times 4$. In the experiment, the images are all up-sampled from $32$ to $224$, and for the band smoothing, the band size $b$ is fixed to $4$ on the original size of $32$. Table~\ref{tab:cifar10} shows clean and certified accuracy against patches of different sizes. Experiments show that ECViT can effectively improve smoothed ViT and achieve state-of-the-art accuracy. For ViT-S, under the same structure, ECViT-S improves clean accuracy by 7.16\% and certified accuracy by $\sim$13\%. Our best ECViT-B still has 82.80\% certified accuracy under $2\times2$ patches with 93.48\% clean accuracy.

\subsection{Certification on ImageNet}
We evaluate our proposed ECViT on ILSVRC2012 validation set~\cite{ilsvrc}. ILSVRC2012 validation set has 50,000 images and we test clean accuracy and certified accuracy on the whole 50,000 images. Following the setting of previous work~\cite{ds,PatchGuard}, we select three different patch sizes, including 1\% ($23 \times 23$), 2\% ($32 \times 32$) and 3\% ($39 \times 39$). Table~\ref{tab:imagenet} shows clean and certified accuracy compared state-of-the-art methods on ILSVRC2012. ECViT-B surpasses BagCert's clean and certified accuracy by 7.48\% and 9.47\% and achieves state-of-the-art clean and certified accuracy in comparison to the previous state-of-the-art methods.

Table~\ref{tab:smoothmodel} shows clean and certified accuracy compared other smooth models on ILSVRC2012. Smooth models compared to other than ECViT are based on different backbones in DS. 
Inference time is calculated when the batch size of the image is 1024 to complete the vote in seconds. To simulate the inference in practice, inference time contains the time of the complete process, including data pre-processing and inference.
In the case of band size $b=37$, our ECViT-B improves the clean accuracy by 3.9\% compared to ViT-B. The certified accuracy is improved by about 10\% in different patch sizes, and it speeds up 4-8x times for ViT-B in the inference time. In the smoothing model, we achieve the fastest inference efficiency. 
Our method achieves state-of-the-art certification on ILSVRC2012 while maintaining 78.58\% clean accuracy, which is very close to the normal ResNet-101. 

\begin{figure}[t]
\begin{minipage}[b]{1\linewidth}
  \centering
  \centerline{\includegraphics[width=8cm]{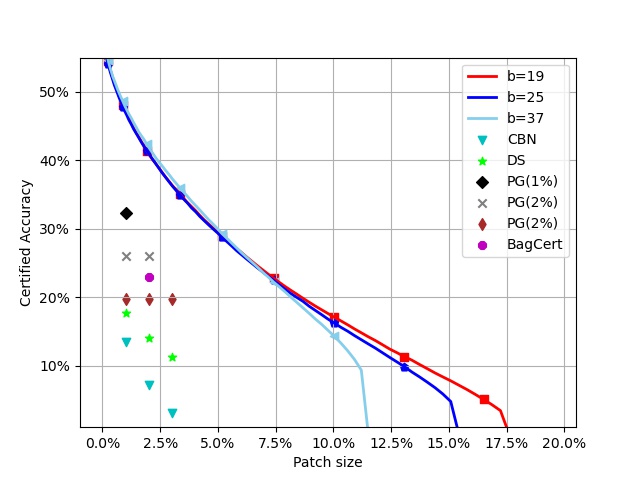}}
\end{minipage}
\caption{Certified accuracy under different patch sizes on ILSVRC2012. Patch size represents the proportion of adversarial patch in the image.}
\label{fig:alb_imagenet}
\end{figure}

\begin{figure}[t]
\begin{minipage}[b]{1\linewidth}
  \centerline{\includegraphics[width=8cm]{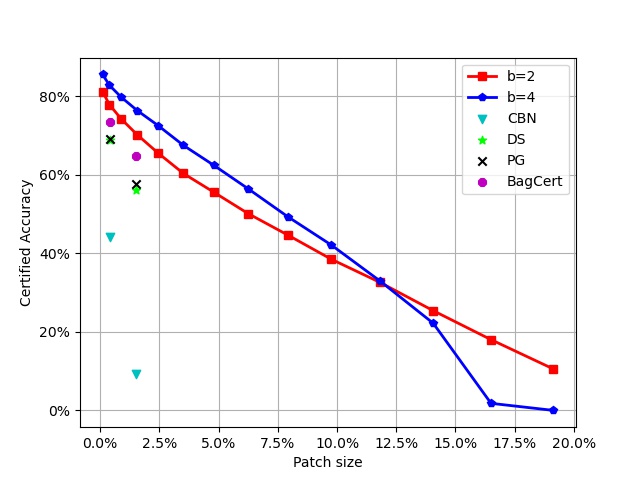}}
\end{minipage}
\caption{Certified accuracy under different patch sizes on CIFAR10. Patch size represents the proportion of adversarial patch in the image.}
\label{fig:alb_cifar}
\end{figure}

\subsection{Ablation Study}
In this section, we mainly focus on studying the effect of the number of stages, different patch sizes, and tokenizers on the clean and certified accuracy.

\noindent\textbf{Number of Stages.} 
To verify the effectiveness of multi-stage progressive smoothed image modeling, we study the effect of different stages on accuracy. Specifically, taking $two\_stage$ as an example, $two\_stage$ represents the direct fine-tuning after the progressive smoothed image modeling of the first two stages. Among them, the total training epoch is the same. 
The meanings of $one\_stage$ and $three\_stage$ are similar with $two\_stage$.
Table~\ref{tab:stage_imageNet} shows the training of different stages and tokenizers on ILSVRC2012 validation set for ECViT-B.
Table~\ref{tab:stage_cifar10} reflects the different stages of ablation experiments on CIFAR10. 
Clean and certified accuracy basically increases with the increase of training stages.
Experiments show that the progressive smoothed image modeling task allows the base classifier to explicitly capture the local context of an image while preserving the global semantic information. Consequently, more discriminative local representations can be obtained through a very limited image information (a smoothed thin image bands), which improves the performance of the base classifier.

\begin{table}[t]
\centering
\caption{Ablation study of training for different stages on CIFAR10. Clean and certified accuracy basically increases with the increase of training stages.}
\label{tab:stage_cifar10}
\scalebox{0.73}{
\begin{tabular}{@{}c|c|c|c|cc@{}}
\toprule
\multirow{2}{*}{Networks} & \multirow{2}{*}{Band Size} & \multirow{2}{*}{Stages} & \multirow{2}{*}{\makecell[c]{Clean\\Accuracy (\%)}} & \multicolumn{2}{c}{Certified Accuracy (\%)} \\ \cmidrule(l){5-6} 
 &  &  &  & \multicolumn{1}{c|}{$\quad 2\times 2 \quad$} & \multicolumn{1}{c}{$4\times4$} \\ \midrule
\multirow{6}{*}{ECViT-S} & \multirow{3}{*}{b=2} & one\_stage & 77.94 & \multicolumn{1}{c|}{66.36} & 57.86 \\ \cmidrule(l){3-6} 
 &  & two\_stage & 77.84 & \multicolumn{1}{c|}{66.56} & 56.80 \\ \cmidrule(l){3-6} 
 &  & three\_stage & \textbf{80.50} & \multicolumn{1}{c|}{\textbf{69.06}} & \textbf{59.58} \\ \cmidrule(l){2-6} 
 & \multirow{3}{*}{b=4} & one\_stage & 86.56 & \multicolumn{1}{c|}{71.26} & 62.60 \\ \cmidrule(l){3-6} 
 &  & two\_stage & 86.30 & \multicolumn{1}{c|}{71.36} & 62.38 \\ \cmidrule(l){3-6} 
 &  & three\_stage & \textbf{87.56} & \multicolumn{1}{c|}{\textbf{73.82}} & \textbf{65.10} \\ \midrule
\multirow{6}{*}{ECViT-B} & \multirow{3}{*}{b=2} & one\_stage & 85.60 & \multicolumn{1}{c|}{76.42} & 68.66 \\ \cmidrule(l){3-6} 
 &  & two\_stage & 86.06 & \multicolumn{1}{c|}{77.20} & 69.06 \\ \cmidrule(l){3-6} 
 &  & three\_stage & \textbf{86.40} & \multicolumn{1}{c|}{\textbf{77.86}} & \textbf{70.20} \\ \cmidrule(l){2-6} 
 & \multirow{3}{*}{b=4} & one\_stage & 92.46 & \multicolumn{1}{c|}{81.54} & 74.56 \\ \cmidrule(l){3-6} 
 &  & two\_stage & 93.14 & \multicolumn{1}{c|}{82.40} & 75.98 \\ \cmidrule(l){3-6} 
 &  & three\_stage & \textbf{93.48} & \multicolumn{1}{c|}{\textbf{82.80}} & \textbf{76.38} \\ \bottomrule
\end{tabular}
}
\end{table}

\noindent\textbf{Patch Sizes.}
The size of adversarial patches will greatly affect the certification. Figure~\ref{fig:alb_imagenet} and Figure~\ref{fig:alb_cifar} respectively reflect the certified accuracy variation on ImageNet and CIFAR10 for ECViT-B. The certified accuracy decreases as the patch size becomes larger. When the patch size reaches 10\%, ECViT-B still has a certified accuracy of $\sim17.00\%$ and $\sim39.00\%$ on ImageNet and CIFAR10.

\noindent\textbf{Tokenizer.}
In order to verify that progressive smoothed image modeling is also effective for other tokenizers, we conduct the following experiments on ECViT-B.
Table~\ref{sota} illustrates the comparison of different tokenizers and other network architectures. We can see that compared to ViT-B, the distilled tokenizer has a significant improvement, but it is still lower than the VAE tokenizer. This also verifies that our method can be adapted to different tokenizers.

\begin{table}[t]
\caption{Clean and certified accuracy compared with other tokenizers on ILSVRC2012. The VAE is better than the distilled.}
\label{sota}
\scalebox{0.7}{
\begin{tabular}{@{}c|c|c|c|c|c@{}}
\toprule
\multirow{2}{*}{Band size} & \multirow{2}{*}{Smooth Model} & \multirow{2}{*}{\begin{tabular}[c]{@{}c@{}}Clean\\ Accuracy (\%)\end{tabular}} & \multicolumn{3}{c}{Certified Accuracy (\%)}     \\ \cmidrule(l){4-6} 
                           &                               &                                                                                 & 1\% pixels     & 2\% pixels     & 3\% pixels     \\ \midrule
\multirow{3}{*}{b=19}      & ViT-B(b=19)                   & 66.92                                                                           & 34.65          & 28.71          & 24.80          \\ \cmidrule(l){2-6} 
                           & Ours(distilled)               & 72.93                                                                           & 45.78          & 40.03          & 35.61          \\ \cmidrule(l){2-6} 
                           & Ours(vae)                     & \textbf{73.49}                                                                  & \textbf{46.83} & \textbf{40.72} & \textbf{36.29} \\ \midrule
\multirow{3}{*}{b=25}      & ViT-B(b=25)                   & 70.57                                                                           & 36.72          & 31.13          & 26.80          \\ \cmidrule(l){2-6} 
                           & Ours(distilled)               & 75.12                                                                           & 46.63          & 40.93          & 36.49          \\ \cmidrule(l){2-6} 
                           & Ours(vae)                     & \textbf{75.30}                                                                  & \textbf{46.56} & \textbf{40.79} & \textbf{36.21} \\ \midrule
\multirow{3}{*}{b=37}      & ViT-B(b=37)                   & 74.68                                                                           & 37.61          & 31.88          & 27.44          \\ \cmidrule(l){2-6} 
                           & Ours(distilled)               & 78.05                                                                           & 46.46          & 40.84          & 36.39          \\ \cmidrule(l){2-6} 
                           & Ours(vae)                     & \textbf{78.58}                                                                  & \textbf{47.39} & \textbf{41.70} & \textbf{37.26} \\ \bottomrule
\end{tabular}
}
\end{table}

\section{Conclusions}
Certifiable patch defenses allow guaranteed robustness against all possible attacks for the given threat model.
Existing certifiable patch defenses sacrifice the clean accuracy of classifiers and only obtain a low certified accuracy on toy datasets, which limits their applications in practice.
To move towards practical certifiable patch defense, we introduce ViT into the framework of DS. With the progressive smoothed image modeling and isolated band unit self-attention, our ECViT obtains state-of-the-art accuracy and efficient inference efficiency on CIFAR-10 and ImageNet. 
In our work, we hope to provide a new perspective on how a suitable network architecture can improve the upper certification of patch defenses in practice.

\noindent\textbf{Limitations and broader impacts.} In spite of the limitation that adversarial patches are assumed to be square in shape rather than rectangular or irregular in shape, we believe that ECViT introduces a practical certifiable patch defense in order to assist physical systems in mitigating the threat of patch attacks.

\section*{Acknowledgements}
This work was supported by National Natural Science Foundation of China (No.62072112), National Key R\&D Program of China (2020AAA0108301), Scientific and Technological Innovation Action Plan of Shanghai Science and Technology Committee (No.20511103102), Fudan University-CIOMP Joint Fund (No. FC2019-005), Double First-class Construction Fund (No. XM03211178).
{\small
\bibliographystyle{ieee_fullname}
\bibliography{egbib}
}

\end{document}